\documentclass{article}

\usepackage{arxiv}

\usepackage[utf8]{inputenc} % allow utf-8 input
\usepackage[T1]{fontenc}    % use 8-bit T1 fonts
\usepackage{hyperref}       % hyperlinks
\usepackage{url}            % simple URL typesetting
\usepackage{booktabs}       % professional-quality tables
\usepackage{amsfonts}       % blackboard math symbols
\usepackage{nicefrac}       % compact symbols for 1/2, etc.
\usepackage{microtype}      % microtypography
\usepackage{lipsum}		% Can be removed after putting your text content
\usepackage{graphicx}
\usepackage{natbib}
\usepackage{doi}
\usepackage{multirow}

\title{The Death of Feature Engineering? \\
—— BERT with Linguistic Features on SQuAD 2.0}

\author {
    Jiawei Li\\
      Department of Energy Resource Engineering\\
      Stanford University\\
      Stanford, CA 94306 \\
      \texttt{jiaweili@stanford.edu}
	\And
      Yue Zhang \\
      Department of Mechanical Engineering\\
      Stanford University\\
      Stanford, CA 94305 \\
      \texttt{yzhang16@stanford.edu} \\
}

\renewcommand{\headeright}{}
\renewcommand{\shorttitle}{The Death of Feature Engineering?
—— BERT with Linguistic Features on SQuAD 2.0}

\begin{document}
\maketitle

\begin{abstract}
Machine reading comprehension is an essential natural language processing task, which takes into a pair of context and query and predicts the corresponding answer to query. In this project, we developed an end-to-end question answering model incorporating BERT and additional linguistic features. We conclude that the BERT base model will be improved by incorporating the features. The EM score and F1 score are improved 2.17 and 2.14 compared with BERT(base). Our best single model reaches EM score 76.55 and F1 score 79.97 in the hidden test set. Our error analysis also shows that the linguistic architecture can help model understand the context better in that it can locate answers that BERT only model predicted "No Answer" wrongly. 
\end{abstract}

% keywords can be removed

\section{Introduction}
Machine reading comprehension has become a central task in natural language understanding. A large amount of model has been explored to this question answering task, including RNN-based model, transformer-based model, Pre-trained Contextual Embeddings model. The recent development of BERT has shown good results on many different Natural Language Processing tasks and achieved good performance. It also works quite well on question answering systems, as manifested by the leaderboard of SQuAD 2.0. Thus, in our project we also use BERT model as the backbone model.

\begin{figure}[htp]
\label{fig:intro}
    \centering
    \includegraphics[width=0.75\textwidth]{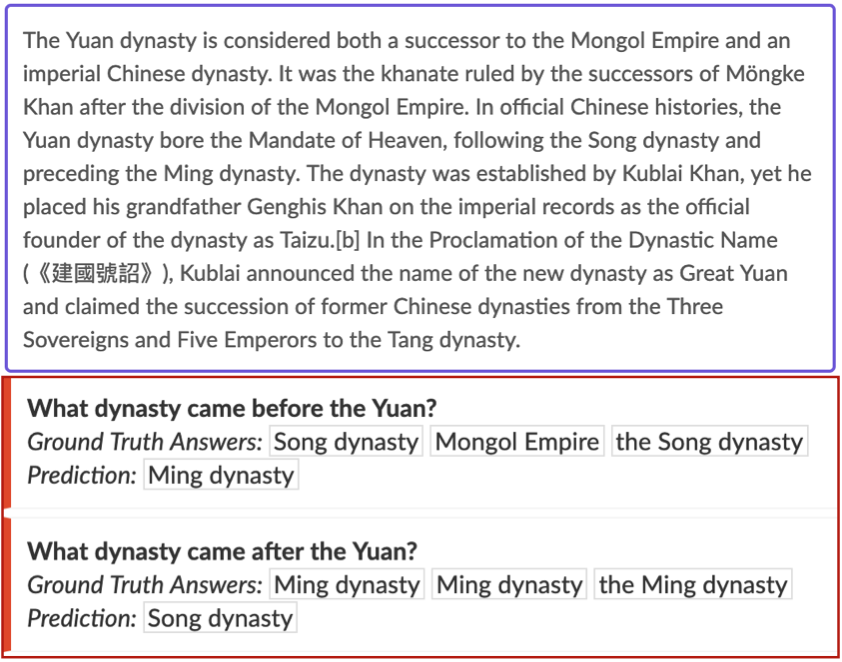}
    \caption{Predictions of BERT (single model) (Google AI Language)~\cite{devlin2018bert} on SQuAD}
\end{figure}

However, despite the high metric score, we still find some basic NLU errors, especially when the linguistic structures are complicated. As shown in figure~\ref{fig:intro}, we can see that BERT model~\cite{devlin2018bert} predict wrong answers encountering complex logic and linguistic structure.

Thus, It motivates us to develop a SQuAD model incorporating add additional linguistic features to help the model understand the linguistic structures better so as to reduce the number of NLU errors in previous BERT model predictions. We leveraged the NLP package SpaCy to extract linguistic features from the context and the question. Our proposed model improved EM and F1 score by 2.17 and 2.14 compared BERT(base) on the dev set. We also employed our method on BERT(large) and reach a EM and F1 score at 76.55 and 79.97 at the final test set. 

We further discussed the errors for certain examples and showed that our model can predict correct answers where BERT predicts no answer when the linguistic structure is complex. We also show that the major errors currently we have is the model inability to determine whether the answer exists for a certain question. The further improvement for SQuAD could be more optimized treatment of no answer situations

\section{Related Work}
RNN based model is first proposed in the reading comprehension tasks. This kind of model \cite{seo2016bidirectional} generally consists of RNN encoders for context and query and RNN modeling layer to predict the answer position, an attention layer is then applied to help query hidden states focus on the certain words. 
 
 However, RNN based model has two significant drawbacks. First, due to its recurrent sequential natural, it is very difficult to perform the parallel computation on the GPU, which restrains its ability to incorporate large scale dataset. Second, it is very diffcult to preserve long term relationships in the paragraphs even though with gated scheme in LSTM or GRU. Thus people began to propose non-RNN models for general NLP tasks. Transformer~\cite{vaswani2017attention} is one of the most successful non-RNN models , outperforms both recurrent and convolutional model in the NMT translation task at the time when it was proposed. General transformer consists of an encoder and a decoder. Each encoder or decoder consists of several transformer block. A typical transformer block consists of multi-head self attention, skip connection and feed forward layers. Transformer is capable of handing large scale data. 
 
 BERT~\cite{devlin2018bert} refers to Bidirectional Encoder Representations from Transformers. It is a very power pre-training contextual embeding language model. BERT can be trained on large scale unlabeled text and then be used to embed the text for a certain nlp task. The general structure of the bert integrate a transformer encoder as its language model. Especially \cite{devlin2018bert} employed a masked language model to train its transformer encoder in the large scale unlabeld text in a directional fashion. Then the pre-trained encoder can be used for downstream nlp tasks. For question answering problem in SQuAD, the input representation is the concatenation of toke embeddings, segment embeddings and position embeddings for both context and query. It has been shown that bert-based model has achieved out-performs other models in the SQuAD leaderboard.

 \section{Approach}
As described above, the BERT model has been successfully used in many different NLP tasks and performs quite well. In terms of Question Answering tasks, it has been dominating the leaderboard for SQuAD 2.0 challenge since emerging and the performance is quite close to human level already. However, there are still some natural language understanding issues that the predictions of BERT model is not able to solve. To address these issues, we motivate our model architecture by incorporating some linguistic features in BERT to help. We now illustrate our architectures below in two parts, BERT model and the additional linguistic layers. The folloing figure shows our architecture clearly.

\begin{figure}[h]
    \centering
    \includegraphics[width=0.85\textwidth]{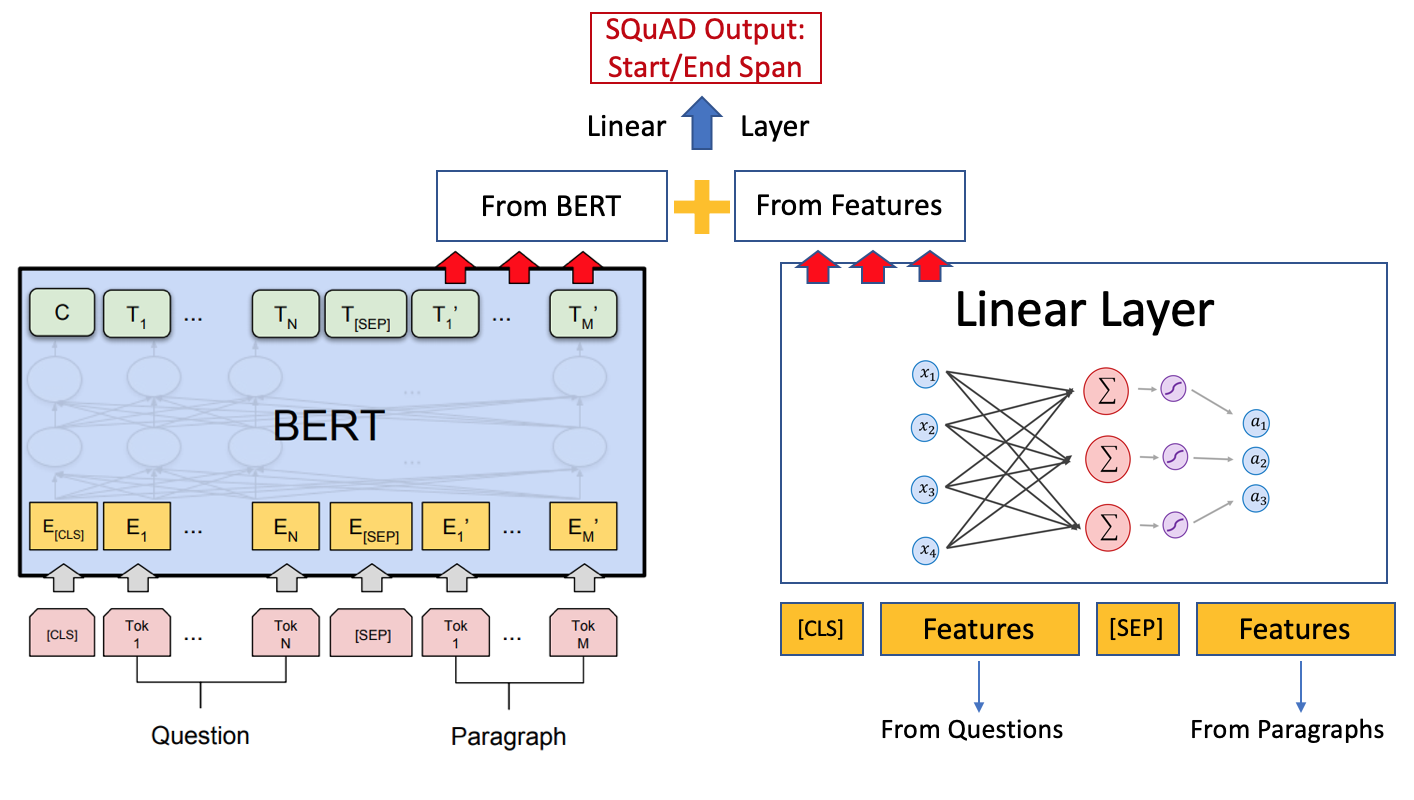}
    \caption{Designed BERT model architecture with linguistic features.}
    \label{fig:my_label}
\end{figure}

\subsection{Backbone model: BERT}
In this project, we use the pre-trained contextual model BERT as our backbone model. The general approach is to use BERT to encode a context and question pair and directly predict the start point and end point for the answer. For each pair of paragraph context and question, the input for BERT consists of three components. The dimension of all the three components are all (batch\_size, max\_seq\_len), which are hyperparameters defined in the model. The first component is the word index for context and questions. Especially, we will add "[CLS]" to represent the start of the context, "[SEP]" to separate the context and the question. The second part is the segmentation input, which basically differentiate the word from context and question. The last component is a mask used identify the padding numbers in the batch. The output of BERT can then be used to predict the start and end position for the answer prediction. However, we do not directly predict the position from BERT but treat the BERT output as part of the input a next linear layer as illustrated in Figure 1. In our implementation of BERT, we adapted some code from huggingface~\cite{bert_implementation}.

% \begin{figure}[h]
%     \centering
%     \includegraphics[width=0.65\textwidth]{architecture.png}
%     \caption{Designed BERT model architecture with linguistic features.}
%     \label{fig:my_label}
% \end{figure}

\subsection{Additional linguistic features layer}
The linguistic features contain token level features for both questions and answers. We choose to use token level features to be consistent with the token level BERT input. Each token has four features below and all these are extracted by using a Python NLP package SpaCy.
\begin{itemize}
    \item NER: Name entity label
    \item POS: The part-of-speech tag
    \item DEP: Syntactic dependency, i.e. the relation between tokens
    \item STOP: Is the token part of a stop list, i.e. the most common words of the language ("is", "the" will be considered in the stop list)
\end{itemize}
The output of NER, POS and DEP are all string labels. For each of the three features, we map the string labels to integers as input. For example, we map 23 different name entity labels to integers from 0 to 22. The last feature STOP will be a 0/1 vector and we don't need further processing. Since each BERT example consists of paragraph context and a question, we extract these four features for tokens in both context and questions and then concatenate them as a vector. We then pad the vector to a size of max\_seq\_len, which is the same as the input size in BERT model. We then pass these additional features to a linear layer followed by a relu layer and the resulting vectors contain information of all these linguistic features.

\subsection{Final layer}
We concatenate the output from the BERT model and the output from the linguistic features linear layer and treat this as the input for the final linear layer. The output for the final linear layer will then be the start/end span.

\section{Experiments}
\subsection{Dataset}
There are many dataset available on question answering systems, but in this project we focus on Stanford Question Answering Dataset 2.0 (SQuAD 2.0). This dataset is an extension of the widely studied SQuAD 1.1. It combines the 100,000 questions in SQuAD 1.1 with over 50,000 new, unanswerable questions written adversarially by crowdworkers to look similar to answerable ones. Each examples consists of a paragraph of context and a question. The answer to every question is a segment of text, or span, from the corresponding reading passage, or the question might be unanswerable~\cite{rajpurkar2018know}. This dataset is harder than SQuAD 1.1 in that model needs also to determine whether an answer exists or not.

\subsection{Evaluation metric(s)}
There are two metrics for evaluating the result, Exact Match (EM) and F1 score. The Exact Match gives zero-one accuracy based on whether the prediction matches exactly one of the given candidate answers. In contrast, F1 score takes system and each gold answer as bag of words and calculate the averaged F1 score. The score on the leader board now is EM (85.082) and F1 (87.615) recently by MSRA. We will compare our result with both the baseline model provided in the starter code (BiDAF performance) and the scores on the leaderboard of SQuAD 2.0.
\subsection{Experimental details}
The BiDAF baseline model is provided in the started code. We use the same parameters provided in the starter code. For the BERT baseline model, we use the same training procedure as~\cite{bert_implementation}. For training, the max sequential length is set as 384, which is the total length contains context and questions as well separate tokens ("[CLS]", "[SPE]"). The max question length is set as 64 and the max answer length is set as 30. The initial learning rate is set as: 3e-5. We compared the baseline bert model and our model for two different BERT model ( 'bert-base-uncased' and 'bert-large-uncased'). Table~\ref{table:train_setting} shows the details settings for different experiments.

\begin{table}[htp]
\centering
\caption{Training Model Settings}
\begin{tabular}{ c   c   c  c  c  c}
\toprule
\cmidrule{1-6}
Case Name & Bert Model &  Batch size & Accum steps & Epoch & GPU\\
\midrule
BERT (base) & bert\_base & 16& 1 & 4& 4 * k80 \\
Our model (base) & bert\_base & 16 & 1 & 4&4 * k80 \\
BERT (large) & bert\_large & 32 & 2&4& 8 * v100  \\
Our model (large) & bert\_large & 32 & 2 & 4&8 * v100 \\
\bottomrule
\end{tabular}\label{table:train_setting}
\end{table}

\subsection{Results}
The experiments are shown in table~\ref{table:results}. The BiDAF model is trained using the starter code provided by TA. It is treated as the baseline for the model. The reported EM and F1 score are 49.07 and 50.29, respectively. The reported EM and F1 score for our baseline BERT (base) single model is 71.59 and 74.72. Our model with linguistic features has improved EM and F1 score by 2.17 and 2.14, reaching to 73.73 and 76.86. The results shows that incorporating the linguistic features into the our reading comprehensive model can improve the performances. However, when we experiment our new architecture on BERT large model, the EM and F1 results are pretty similar, though both results increased a lot. One of the postulated reason is that BERT base is relatively less powerful in comparison with BERT large and thus adding additional features may see more significant improvement. However, the BERT large model has much more parameters and is already powerful enough to perform very well and thus the additional features may not contribute much. However, training BERT large is really expensive compared to BERT base.

\begin{table}[t]
  \caption{Experiments Results for Single Model}
  \label{table:results}
  \centering
  \begin{tabular}{ccc}
    \toprule
    % \multicolumn{2}{c}{Part}                   \\
    \cmidrule{1-3}
    Model Name     & Dev Set     & Test Set \\
    \midrule
    \textit{Single Model }    &  \textbf{ EM/F1 }    &\textbf{ EM/F1}  \\
    BiDAF  & 49.07/50.29  &  -/-    \\
    BERT (base)     & 71.59/74.72      & -/-   \\
    Our model (base)     & 73.76/76.86       & -/-  \\
    BERT (large)     & 78.51/81.34       & -/-   \\
    Our model (large)    & 78.17/81.20       & \textbf{76.55/79.97}  \\
    \bottomrule
  \end{tabular}
\end{table}

Worth mentioning is that our reported performance are all on single model without ensemble models. We are well-aware that doing ensemble models can further improve our model performance by 1-2 F1 score. However, we did not perform that task because we are more interested in analyzing and interpreting the model and do error-analysis.

\section{Error Analysis}
In this section, we are going to discuss some examples to show how the baseline model fails and how our model performances. As we have explored the prediction results of BERT (single model) (Google AI Language)~\cite{devlin2018bert}, we found that the majority of the errors come from the wrong prediction of Answer/No answer, i.e. the model predicts No answer when there is actually exists answers or the model predicts a certain answer where no answer exists for this certain question. The results are based on our model (large) and BERT (single model) (Google AI Language) in the SQuAD webcite.

\subsection{No Answer Situation}

First, we are going to discuss for two certain examples where the BERT model predicts No answer where there exists answers in table~\ref{table:error}. Our model actually predicts the correct answer in this situation. Analyzing these questions, we can see that they are usually very complex in logic and clause structure, thus the baseline BERT model has difficulties analyzing such sentences with highly complex linguistic structures. However, when we add the linguistic features, i.e name entity, part-of-speech and syntactic dependency, our model can predict the correct answers. The results illustrated that by incorporating the linguistic features into the existing reading comprehensive model based on BERT, it can understand the context better.

This phenomena can be understood quite intuitively. The additional features for both context and questions will do the model a favor on locating the answer span. For example, when the question asks for certain object (What question) or places (Where question), the neural network wants to search for a noun or a place. The feature part-of-speech tags or name entity will be very helpful then. Also, the STOP feature will help the neural network concentrate more on words that really convey information.

\begin{table}[htp]
\centering
\label{table:error}
\begin{tabular}{|c|c|c|c|}
\hline
\multicolumn{4}{|c|}{\textit{Context Part: Yuan dynasty}}                                                                                                                                                                                                                              \\ \hline
\textbf{Question}     & \begin{tabular}[c]{@{}c@{}}What non-Chinese \\ empire did the Yuan \\ dynasty succeed?\end{tabular} & \begin{tabular}[c]{@{}c@{}}What writing inspired \\ the name Great Yuan?\end{tabular} & \begin{tabular}[c]{@{}c@{}}Which tribes did \\ Genghis Khan \\ fight against?\end{tabular} \\ \hline
\textbf{Reference}    & Mongol Empire                                                                                       & \begin{tabular}[c]{@{}c@{}}the Commentaries on\\ the Classic of Changes\end{tabular}  & No Answer                                                                                  \\ \hline
\textbf{BERT-feature} & Mongol Empire                                                                                       & \begin{tabular}[c]{@{}c@{}}the Commentaries on\\ the Classic of Changes\end{tabular}  & Mongol and Turkic                                                                          \\ \hline
\textbf{BERT}         & No Answer                                                                                           & No Answer                                                                             & Mongol and Turkic                                                                          \\ \hline
\end{tabular}
\end{table}

\subsection{Prediction Answers for No-answer Questions}
Next, we focus on another major errors that the baseline BERT model made. In this situation, the ground truth of a certain question is no answer, while BERT base model predicts some answers. The results are shown at the last column in table~\ref{table:error}. We can see that in this situation, both our model and the baseline BERT model predicts wrong answer for a question where no answer actually exists.

\subsection{Confusion Matrix for Prediction No Answers}
We also calculate the confusion matrix for our model to see whether it can successfully predict whether a question has a answer or not. The results are shown in table~\ref{table:conf_matrix}. It is clear shown that our current model is unable to effectively make decisions whether there is an answer or not, even though we have reached a fair high metric on EM/F1. It seems that the major issues for SQuAD 2.0 is to how to correctly deal with no answer situation. Currently, we just manually denote the no answer label as -1, which could be optimized later

\begin{table}[htp]
 \caption{Confusion matrix}
 \centering
 \label{table:conf_matrix}
% \begin{adjustbox}{width=0.65\columnwidth,center}
\begin{tabular}{l|l|c|c|c}
\multicolumn{2}{c}{}&\multicolumn{2}{c}{Predictions}&\\
\cline{3-4}
\multicolumn{2}{c|}{}&Answer&No Answer\\
\cline{2-4}
\multirow{2}{*}{Label}& Answer & 1456 & 1454 \\
\cline{2-4}
& No Answer & 1556 & 1612 \\
\cline{2-4}
\end{tabular}
% \end{adjustbox}
\end{table}

\section{Progress and Future Work}
In our project, we extended the BERT model on Question Answering Dataset SQuAD 2.0 by adding additional linguistic features to help the model to do answer span predictions. The results are quite promising as we get the model performance increased by 2 F1 score on BERT base model. Even though BERT large is super powerful that incorporating additional features did not contribute that much, we still conclude that feature engineering is useful. The reason is that training and using BERT large model is so expensive that it is very hard for practical use in real-world. This shows that feature engineering will not die at least before computational resources become cheaper.

We did error analysis and compared our new model performance with the BERT only model. We found that our model did a better job than the BERT model especially when the linguistic structures of the context and questions are complex. The BERT model may predict "No Answer" under these cases while our model predicts the answer correctly. This shows that the additional linguistic features do help the model to understand the context better and thus is able to locate the correct answer.

More analysis on both BERT and our BERT model with additional linguistic features show that the model is not doing well enough yet on predicting whether a question has answer or not. This may have something to do with the loss function definition and the structure of the neural network architecture, we are interested in exploring this more after the course.

\bibliographystyle{unsrtnat}
\bibliography{references}  %%% Uncomment this line and comment out the ``thebibliography'' section below to use the external .bib file (using bibtex) .





% \end{thebibliography}

\end{document}